\documentclass[runningheads]{llncs}
\usepackage[T1]{fontenc}
\usepackage{graphicx}
\usepackage{amsmath}
\usepackage{amssymb}
\begin{document}
\title{Evaluation of Safety Constraints in Autonomous Navigation with Deep Reinforcement Learning
\thanks{This is a preprint version of the extended abstract accepted to TAROS'23.}
}
%
%
\author{Brian Angulo\inst{1} \and Gregory Gorbov\inst{1,3} \and Aleksandr Panov\inst{2,3} \and Konstantin Yakovlev\inst{2, 3}}
%
%
\institute{Moscow Institute of Physics and Technology, Dolgoprudny, Russia
\and
AIRI, Moscow, Russia
\and
Federal Research Center for Computer Science and Control of Russian Academy of Sciences, Moscow, Russia 
}
%
\maketitle              
\begin{abstract}

While reinforcement learning algorithms have had great success in the field of autonomous navigation, they cannot be straightforwardly applied to the real autonomous systems without considering the safety constraints. The later are crucial to avoid unsafe behaviors of the autonomous vehicle on the road. To highlight the importance of these constraints, in this study, we compare two learnable navigation policies: safe and unsafe. The safe policy takes the constraints into account, while the other does not. We show that the safe policy is able to generate trajectories with more clearance (distance to the obstacles) and makes less collisions while training without sacrificing the overall performance.

\end{abstract}
\section{Introduction}

Deep Reinforcement Learning has demonstrated tremendous success in many high-dimensional control problems, including autonomous navigation. Within RL, the interaction of the agent with the environment is modeled as a Markov decision process (MDP)~\cite{sutton2018reinforcement}, where the goal is to optimize the expected cumulative reward. The agent in MDP has a big freedom to explore any behavior which could improve its performance, including those that might cause damage. To this end, it is crucial to ensure safety constraints. A well-known approach to consider safety constraints in RL is a Constrained Markov Decision Process (CMDP)~\cite{altman1999constrained}
. A survey of methods for solving CMDP can be found in~\cite{liu2021policy}. In this short paper, we will apply safety constraints to ensure the safe behavior of the autonomous vehicle. Particularly, we will use the Lagrangian method~\cite{chow2017risk} that is one of the most widely used approaches for solving CMDP.

\section{Problem Statement}

We are interested in algorithms for autonomous navigation which provide certain safety guarantees. To this end, we model our problem as a CMDP, where the agent (autonomous vehicle) must generate a sequence of actions (trajectory) that drives it to a goal while avoiding obstacles and ensuring a tolerable safety cost limit. The latter in our work is interpreted as an upper limit of the vehicle's velocity when it is moving near to the obstacles.

\textbf{Trajectory generation}  We are interested in autonomous vehicles whose dynamics is described as~\cite{surveyMotionPlanning}: $\dot{x} = v cos(\theta)\nonumber, \dot{y} = v sin(\theta), \dot{\theta} = \frac{v}{L} \tan(\gamma)$, where $x$,$y$ are the coordinates of the vehicle's rear axle, $\theta$ is the orientation, $L$ is the wheel-base, $v$ is the linear velocity, $\gamma$ is the steering angle. The former three variables comprise the state vector: $\boldsymbol{x}(t)=(x,y,\theta)$.
The latter two variables form the control vector: $\boldsymbol{u}(t) = (v, \gamma)$, which can also be re-written using the acceleration $a$ and the rotation rate $\omega$ as follows: $v = v_0 + a \cdot t, \gamma = \gamma_0 + \omega \cdot t$.

The robot is operating in the 2D workspace populated with static obstacles. Their shapes are rectangular (as the one of the robot). Let $Obs$ denote the set of obstacles. Denote by $\mathcal{X}_{free}$ all the configurations of the robot which are not in collision with the obstacles. The problem is to find the actions that move the vehicle from its start configuration $s_{start}$ to the goal one $s_{goal}$, s.t. that the kinodynamic constraints are met and the resultant trajectory lies in $\mathcal{X}_{free}$. These controls are generated using a sequential decision making based on CMDP.

\textbf{Constrained Markov Decision Process} Formally, CMDP can be represented as a tuple $(\mathcal{S}, \mathcal{A}, \mathcal{P}, \mathcal{R}, \mathcal{C}, d, \gamma)$, where $\mathcal{S}$ is the state space, $\mathcal{A}$ is the action space, $\mathcal{P}$ is the state-transition model, $\mathcal{R}$ is the reward function, $\mathcal{C}$ is a constraint cost function and $\gamma$ is the discounting factor. During learning at each time step the agent being in a state $s_t \in \mathcal{S}$ takes an action $a_t \in \mathcal{A}$ and receives a reward $r_t \in \mathcal{R}$ and a cost $c_t \in \mathcal{C}$.  The goal is to learn a policy, i.e. the mapping from the states to the distributions of actions, $\pi: \mathcal{S} \rightarrow P(\mathcal A)$. The policy should maximize the expected return $J(\pi)$ from the start state $s_t$ while satisfying the discounted cost with tolerable limit $d_i$ trough the discounted cost $C(\pi)$ under policy $\pi$:
\begin{equation*}
   J(\pi) = \mathbb E_{\tau \sim \pi}[\sum_{i=0}^T \gamma^{t}r(s_t, a_t, s_{t+1})],\;\; C(\pi) = \mathbb E_{\tau \sim \pi}[\sum_{t=0}^T \gamma^{t}c_i(s_t, a_t, s_{t+1})],
\end{equation*}
where $\tau = (s_0, a_0, s_1, a_1, \ldots)$ denotes a trajectory.
The objective of CMDP for the policy $\pi$ is to find: $\pi^* = \arg\max_{\pi \in \Pi} J(\pi), \;\; s.t. \; C_i(\pi) \leq d_i$.
 
\section{Method}

To evaluate the influence of safety constraints, we will use a policy-gradient algorithm PPO~\cite{PPO} and its safe version -- LagrangianPPO (LPPO)~\cite{chow2017risk}. The learning and evaluation of these two algorithms are conducted in a gym environment from \cite{angulo2022policy}. Next, we will briefly introduce some details of environment.

\begin{figure*}[t]
    \centering
        \includegraphics[width=0.25\textwidth]{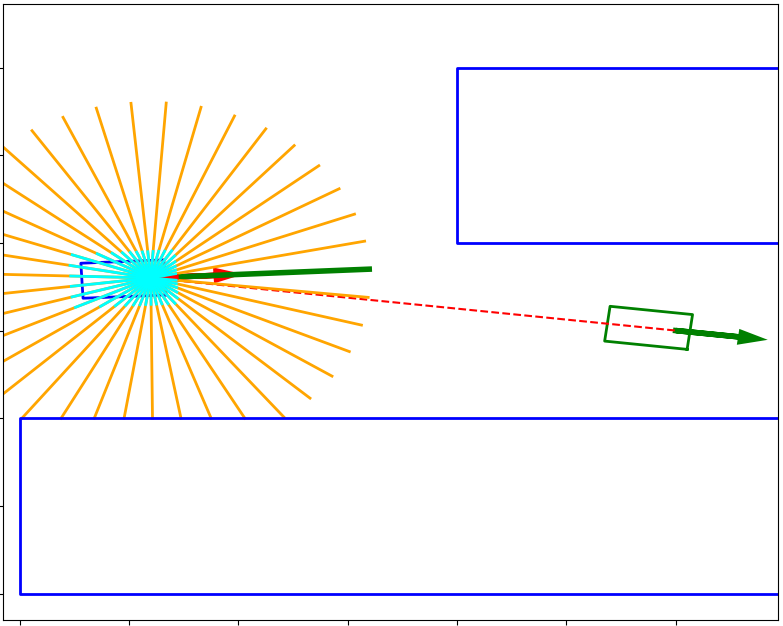}
        \includegraphics[width=0.25\textwidth]{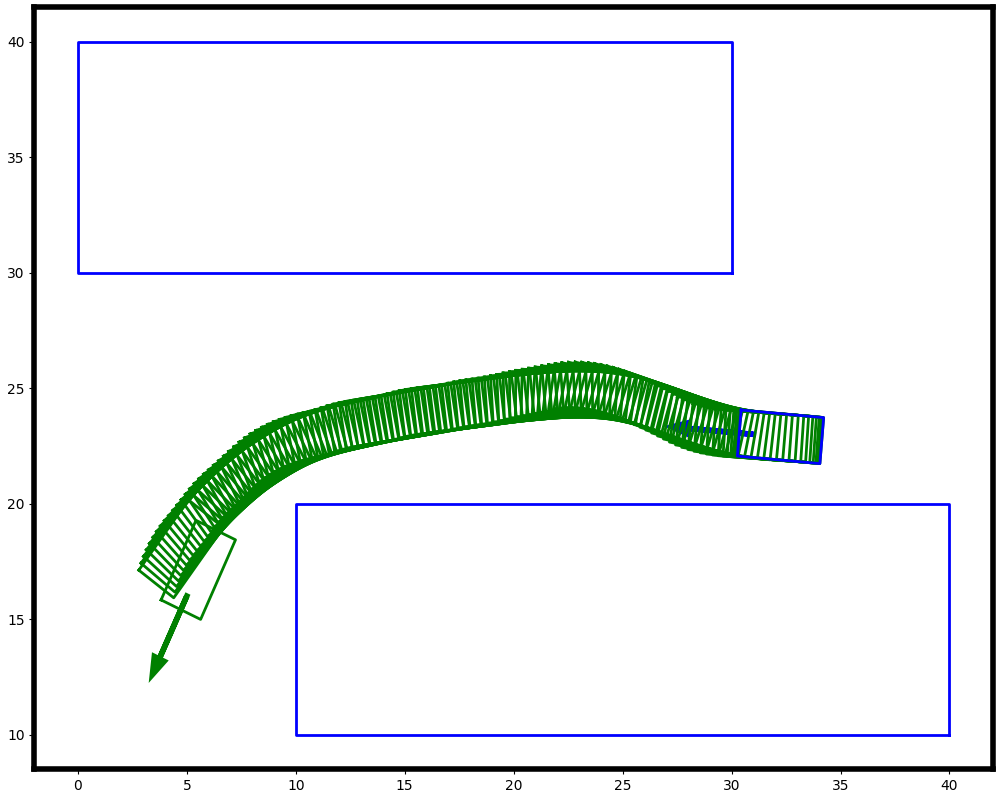}
        \includegraphics[width=0.25\textwidth]{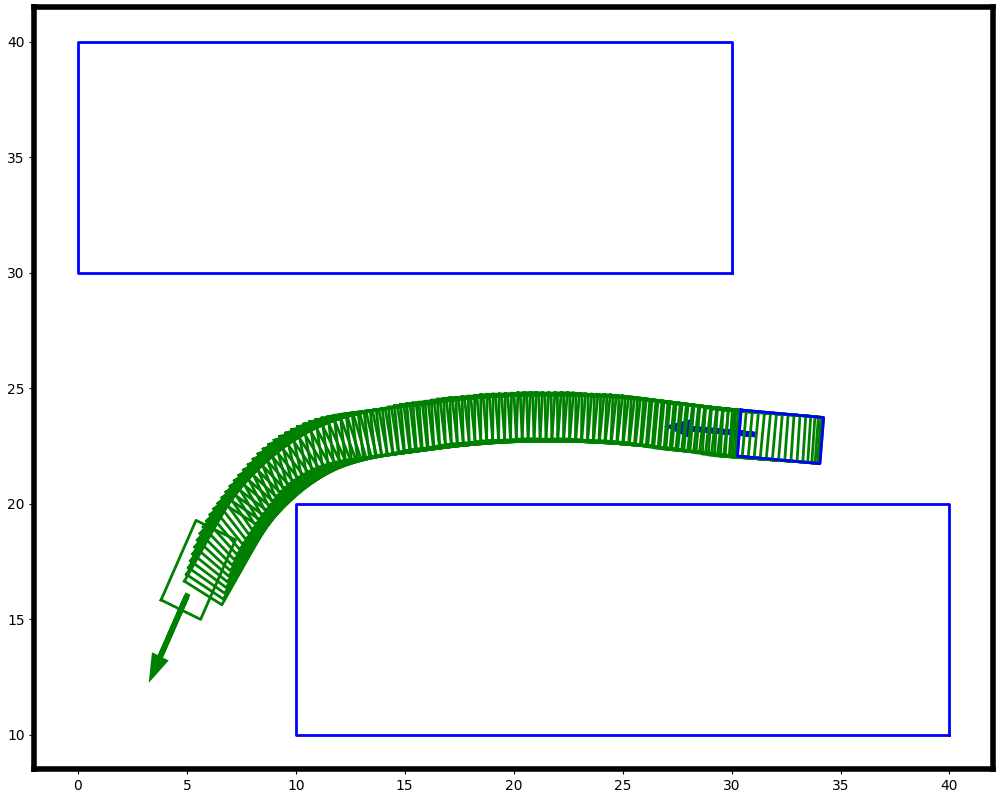}
    \caption{The learning environment is shown \textbf{in the left figure}. \textbf{Blue} rectangle with \textbf{Red} arrow is the current state of the robot. \textbf{Green} rectangle is the goal desired state. \textbf{Blue} rectangles are the static obstacles. \textbf{Orange} lines are the lidar rays and \textbf{Cyan} lines are the safety radius. \textbf{The center and right figures} represent the trajectories generated by LPPO (safe) and PPO (unsafe) policy for the same task respectively.}
    \label{fig:polamp_environment}
\end{figure*}

\textbf{Environment} Our environment provides an autonomous vehicle equipped with kinematic bicycle model and lidar-sensor in an environment which is populated with static obstacles as shown in Fig.~\ref{fig:polamp_environment}. The vehicle state is described as a tuple $\boldsymbol{x} = (x, y, \theta, v, \gamma)$. We consider agent's state $s_t = (\Delta x, \Delta y, \Delta\theta, \Delta v, \Delta\gamma, \theta, v, \gamma, \boldsymbol{l})$ where $\Delta x_{j}$ is the difference between the j-element of the tuple between goal and current vehicle state, $\theta, v, \gamma$ are elements of current vehicle state and $\boldsymbol{l}$ is a tuple of rays-measurements from lidar. In this paper, we consider the same actions and reward function from~\cite{angulo2022policy}. The actions $a_t = (a, \omega)$ are composed of the linear acceleration $a$ and rotation rate $\omega$. For details of the actions, kinematic model and reward function we refer the readers to ~\cite{angulo2022policy}.

\textbf{Safety Constraints}
The safety constraint cost is induced by a velocity of the autonomous vehicle at which it moves near to the obstacles due to the risk of collision like the method proposed in \cite{chow2019lyapunov}. We impose an immediate constraint cost as $c_i(s_i) = \|v_i\| \times 1\{\|\boldsymbol{l}\| \leq r_{safety}\}$, where $r_{safety} = 0.5 m$ is the safety radius -- see Fig.~\ref{fig:polamp_environment}, $\boldsymbol{l}$ is the current lidar signal, and $1$ is boolean function. This safety constraint allows to the agent to move near obstacles at tolerable velocity $v_i$ so as not to violate the tolerable limit $d_i$.

\section{Evaluation}

\begin{figure*}[t]
    \centering
        \includegraphics[width=0.48\textwidth]{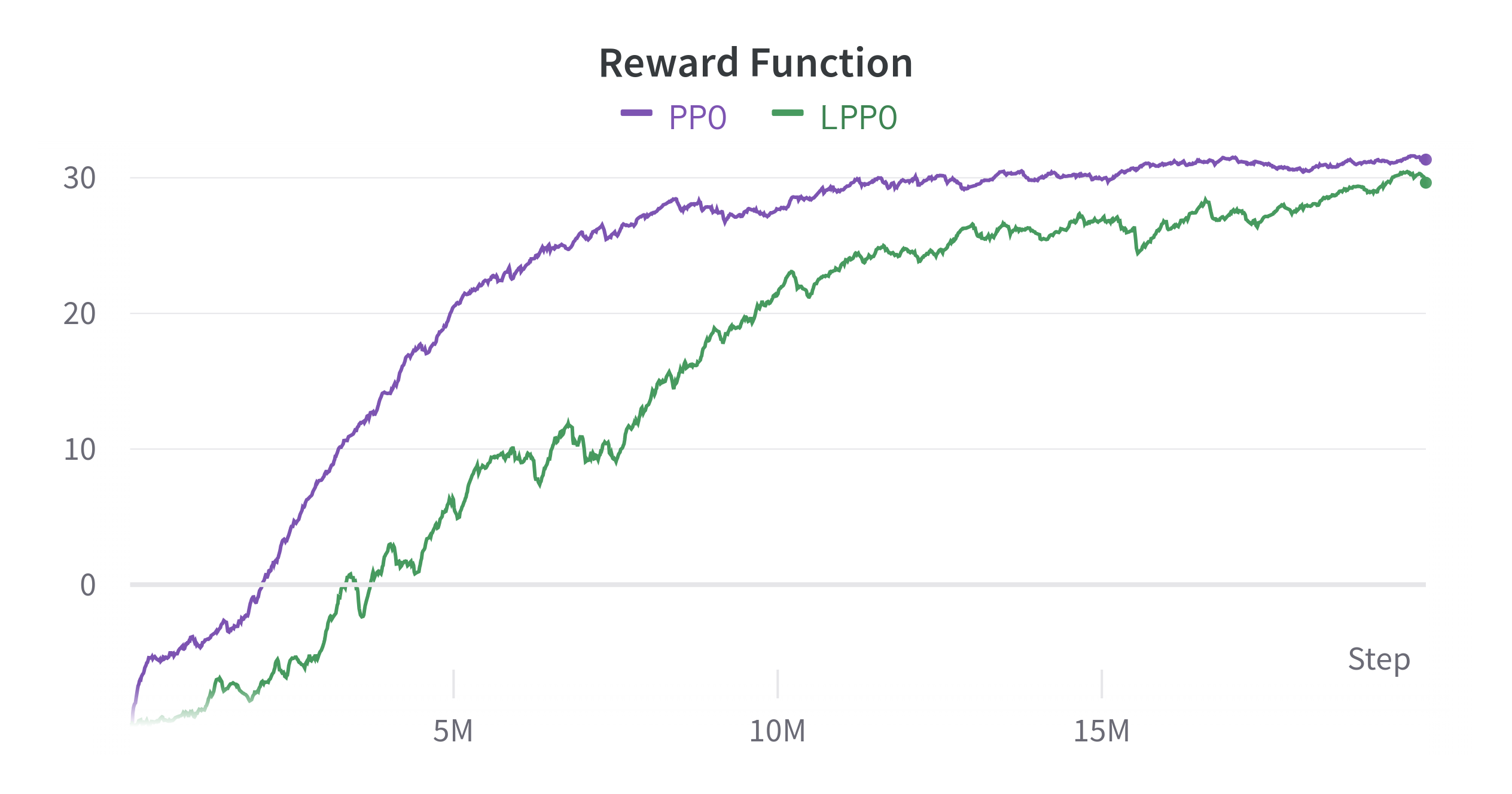}
        \includegraphics[width=0.48\textwidth]{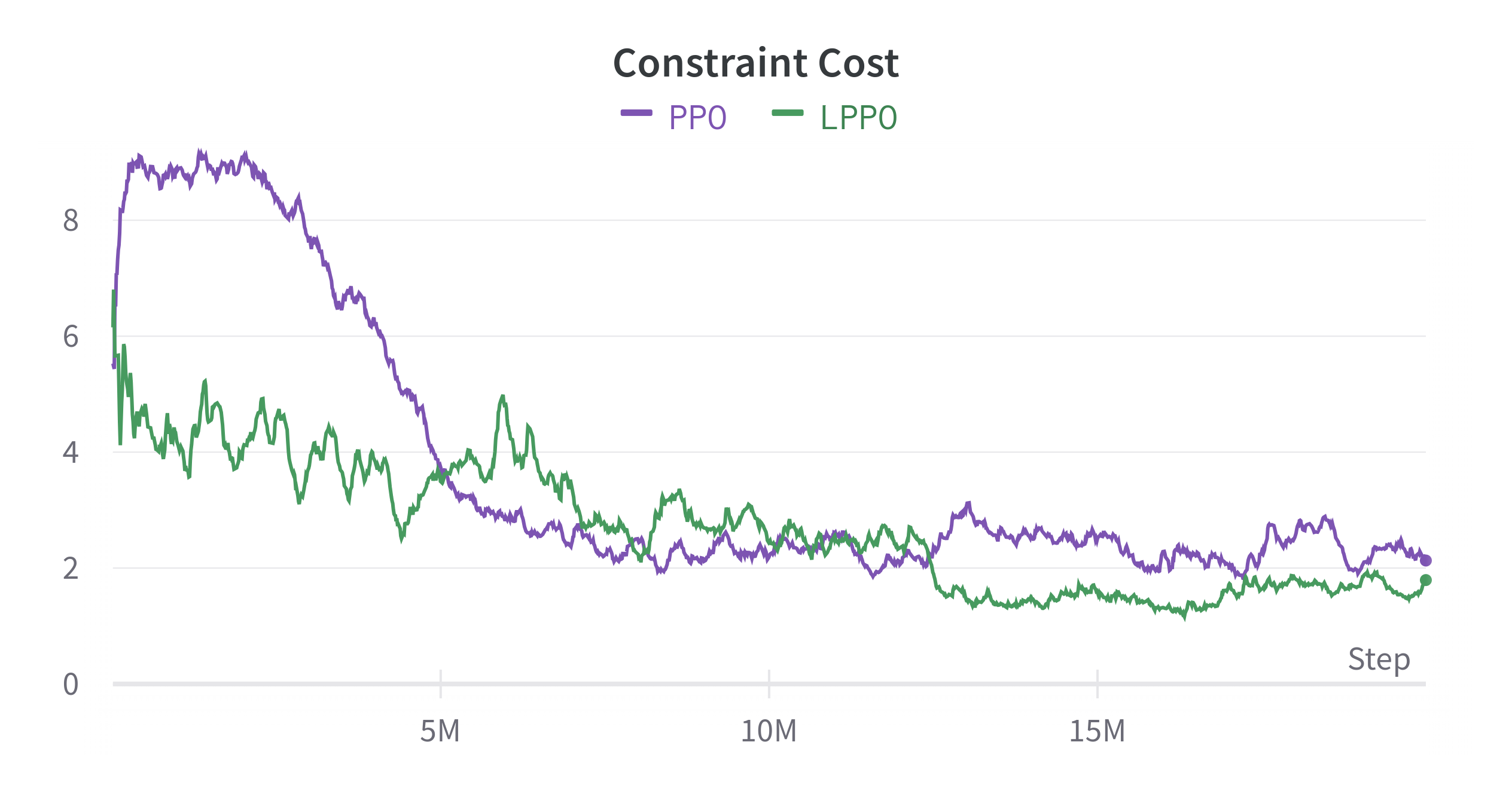}
    \caption{A comparison of learning curves between PPO and LPPO.}
    \label{fig:learning}
\end{figure*}

We trained two policies, i.e. PPO~\cite{PPO} and LPPO~\cite{chow2017risk} on a dataset from \cite{angulo2022policy}. The dataset is comprised of tasks which consist of start, goal and a set of static obstacles. The goal is to generate a trajectory from start to goal. In the Fig.~\ref{fig:learning} we show the learning process of these algorithms. We can see that PPO starts to converge earlier in comparison with LPPO. But at the same time it violates more safety constraints. On the other hand, the LPPO agent tries to find a trade-off.

After training we evaluate these two policies on a validation dataset through the following metrics: success rate (SR, \%), collision rate (CR, \%) and mean minimum clearance distance (MMC, m) of the successfully generated trajectories. The results for PPO are $SR=96.25\%$, $CR = 3.75\%$, $MMC=0.61 m$, and for LPPO are $SR=94.5\%$, $CR = 3.5\%$, $MMC =0.67 m$. We see that the LPPO agent tends to avoid the obstacles with more clearance in order to guarantee the safety constraints as shown in the Fig.~\ref{fig:polamp_environment}. On the other hand, the PPO agent acquired more aggressive behavior that tries to reach the goal as soon as possible regardless of the risk of collision. Additionally, we can see that LPPO agent finishes some tasks without success or collision. In this cases, the LPPO agent prefers to make an abrupt stop instead of colliding with an obstacle. The latter is a safe behavior that agent acquired thanks to the safety constraints.

\section{Conclusion}

In this short paper, we have evaluated the consideration of safety constraints during the optimization process for policy gradient algorithm PPO. While PPO can converge earlier and has a slightly high success rate, the learned behavior of LPPO is more conservative due to taking the safety constraints into account. Overall, considering safety constraints in the optimization problem provides a more conservative and safe behavior without sacrificing of performance.

%
%
%
%
\bibliographystyle{IEEEtran}
\bibliography{root}

\end{document}